# DBC based Face Recognition using DWT

Jagadeesh H S[1], Suresh Babu K[2], and Raja K B[2]

[1]Dept. of Electronics and Communication Engineering, A P S College of Engineering, Bangalore, India
*Email: jagadeesh.11@gmail.com*
[2]Dept. of Electronics and Communication Engineering, University Visvesvaraya College of Engineering, Bangalore University, Bangalore, India

**ABSTRACT**

*The applications using face biometric has proved its reliability in last decade. In this paper, we propose DBC based Face Recognition using DWT (DBC- FR) model. The Poly-U Near Infra Red (NIR) database images are scanned and cropped to get only the face part in pre-processing. The face part is resized to 100\*100 and DWT is applied to derive LL, LH, HL and HH subbands. The LL subband of size 50\*50 is converted into 100 cells with 5\*5 dimention of each cell. The Directional Binary Code (DBC) is applied on each 5\*5 cell to derive 100 features. The Euclidian distance measure is used to compare the features of test image and database images. The proposed algorithm render better percentage recognition rate compared to the existing algorithm.*

## 1. INTRODUCTION

Smart automatic digital systems development is posed to varying constraints with time. The man machine interface decides the smartness of any designs, depending on the percentage of human intervention required in any application. Biometrics is one such area providing smarter solutions to real time problems. The human organs with or without traits are considered as parameters of authentication and or verification purpose. The different biometric parameters include fingerprint, palm print, iris, DNA, typing rhythm, speech, face and others. Recent additions to the list are brain, age, gender, height, veins, lip motion, smile [1], teeth, gesture, emotion, race, attitude and open for add-ons. The real time situations cannot be predicted and depicted, since humans also make false decisions. The development of efficient and accurate systems design has culminated with confined environments. The robust and efficient system requirement in all respects is always persistent. Better results are achieved with multiple biometric parameters for complex systems such as dealing with huge and quality less input data. Many systems accept image as input data.

The major steps involved in a general recognition system are; image acquisition, pre-processing, feature extraction and comparison. A camera acquires good quality image and refinement is done in pre-processing. In feature extraction, morphological or textual attributes are elicited, which uniquely represents the input image. Some of the tools used for feature extraction are Principal Component Analysis (PCA) [2], Discrete Cosine Transform (DCT) [3], Discrete Wavelet Transform (DWT) [4] and Local Binary Patterns (LBP) [5]. The LBP extracts the relation between given pixel with its neighbouring pixels. The major disadvantage of LBP is it does not provide directional information of the image. The Directional Binary Codes (DBC) [6] retains all the advantages of LBP by overcoming the disadvantage of the same. Matching between stored attributes of database and a test image is made in comparison to give the decision of status of affiliation. The recognition rate depends on input image quality, number of subjects and tools used in all steps.

The FR is preferred over iris and fingerprint, because it does not require cooperation of the user and is non intrusive. If the FR System fails for a valid user, it is verified through the naked eye also, which is very difficult in other systems. The performance of any algorithm is influenced by illumination, pose, expression, occlusion, hairstyle, makeup, background, beard, eyeglasses and many other parameters. The FacE REcognition Technology (FERET), Face Recognition Vendor Test (FVRT) 2000, 2002, 2006, Face Recognition Grand Challenge (FRGC) v1.0, FRGC v2.0, Motion Pictures Experts Group-7 (MPEG-7), Expanded Multi Modal Verification for Teleservices and Security (XM2VTS) are some evaluation protocols. Extended Yale-B, CAS-PEAL-R1, Olivetti and Oracle Laboratory (ORL), Indian database, Face recognition data (University of Essex, UK), AR face database and many other databases are used for the evaluation of algorithms.  Each database contains images with the variation of one or more parameters such as illumination, facial expression and others, acquired over a period of time. The number of images in a database range from hundreds to thousands, which are proportional to the accuracy of recognition. The surveillance, ATMs for bank, driving licenses, passports, home town security, law enforcement [7], access control, tele-medicine, voter identification and any restricted entry confidential work applications use the face recognition.

*Contribution:* In this paper DBCFR algorithm is proposed to identify a person. The image is pre-processed and DWT is applied. The DBC technique is used to extract features from LL sub band only. The Euclidian distance is used for the comparison between test image and database images.

*Organization:* Rest of the paper is organized into following sections; section 2 is an overview of related work. The DBCFR model is described in section 3. Section 4 explains the algorithm used in entire model. Performance analysis is discussed in section 5 and conclusion is given in section 6.

## 2. LITERATURE SURVEY

Zhao et al., [8] conducted survey on face recognition as it has received significant attention in recent years. After few decades of research it is able to provide promising solutions for the applications such as commercial and law enforcement. The availability of protocols viz., FERET, XM2VTS, and MPEG-7 has made researchers to improve algorithms with constraints. Tahia Fahrin Karim et al., [9] implemented a reliable PCA based face recognition system and evaluated the performance using standard face databases such as Indian database and the Face recognition data, University of Essex, UK**.** The different techniques such as sum of absolute difference, sum of squared difference and normalized cross correlation are used for matching unknown images with known images.

Gopa Bhaumik et al., [10] proposed an efficient algorithm to detect human behaviours for visual surveillance. The face recognition technique is for dynamic scenario using PCA and minimum distance classifier. The mathematical analysis is made on the video of human face captured to understand or interpret the behaviour. Sumathi and RaniHema Malini [11] have chosen an average-half-face for facial feature extraction using wavelets and PCA for healthcare application. The optical sensor using the videophone system is a promising biometrics system for E-Health, e.g. telemedicine. The symmetry property of the face is effectively utilized. The substantial savings in storage and computation time is the advantage. Haijun Zhang et al., [12] presented a Two-Dimensional Neighborhood Preserving Projection (2DNPP) for appearance based face representation and recognition. 2DNPP enables us to directly use a feature input of 2D image matrices rather than 1D vector. The results corroborate that 2DNPP outperforms the standard neighbourhood preserving projection approach across all experiments with respect to recognition rate and training time on ORL, UMIST and AR face datasets.

Loris Nanni et al., [13] conducted survey on local binary patterns. First several different approaches are compared, and then the best fusion approach is tested on different datasets and compared with several approaches. The experiments show that a fusion approach based on uniform Local Quinary Pattern (LQP) and a rotation invariant LQP, where a bin selection based on variance is performed and Neighborhood Preserving Embedding (NPE) feature transform is applied to obtain better results on six different datasets with support vector machine classifier. Francisco A. Pujol et al., [14] proposed Principal Local Binary Patterns (PLBP) for recognizing faces. The attribute evaluator algorithm of the data mining tool Weka is used. It is assumed that each face region has a different influence on the recognition process; a 9-region mask is designed and a set of optimized weights are obtained for the mask by means of the data mining tool RapidMiner. The method is tested with the FERET database.

Xiaoyang Tan and Bill Triggs [15] proposed a reliable practical face recognition system under uncontrolled lighting conditions. The three main contributions involve: (i) the relationship between image normalization and feature sets. (ii) robust feature sets and feature comparison strategies (iii) the robustness is improved by adding kernel PCA feature extraction and incorporating rich local appearance cues from two complementary sources—gabor wavelets and LBP—showing that the combination is considerably more accurate than either feature set alone. Mayank Agarwal et al., [16] presented a methodology for face recognition based on information theory approach of coding and decoding the face image. It has two stages – feature extraction using PCA and recognition with feed forward back propagation neural network. The algorithm has been tested on 400 images (40 classes). A recognition score for test lot is calculated by considering almost all the variants of feature extraction.

Ramesha et al., [17] proposed an advanced face, gender and age recognition algorithm that yield good results when only small training set is available and it works even with a training set as small as one image per person. The process is divided into three phases: pre-processing, feature extraction and classification. The geometric features from a facial image are obtained based on the symmetry of human faces and the variation of gray levels, the positions of eyes; nose and mouth are located by applying the canny edge operator. The gender and age are classified based on shape and texture information using posteriori class probability and artificial neural network respectively. Meihua Wang et al., [18] proposed a combined feature extraction method which is based on DWT and DCT for face recognition. First the original face image is decomposed by 2D-DWT, then 2D-DCT is applied to the low frequency approximation image component. The SVM classifier is built, using the DCT coefficient, and face image is recognized. The evaluation on ORL database yields higher recognition rate than the traditional PCA algorithm.

Chin-Shyurng Fahn and Yu-Ta Lin [19] developed an automatic real-time face tracking system installed on a robot, which is able to locate human faces in image sequences from the robot situated in real environments. The robot system interacts with human being properly and makes them to react more like mankind. The particle filter uses fewer samples to maintain multiple hypotheses than conventional filters do, so the computational cost is greatly decreased and performs well for non-linear and non-gaussian problems. Li Meng and Dong Wei [20] proposed a portable robotic system for human face recognition, using basic techniques of image processing, such as different edge detection techniques and harmony variation. It comprises of human interface, data processing and robotic control mechanism. A novel display platform has been built, through testing and the interface is easy to manipulate. The skin area is extracted from the normalized face image. The mouth region, eyebrows and eyes could also be identified by the method used.

Unsang Park and Anil K Jain [21] proposed an automatic facial mark detection method which uses: (i) the active appearance model for locating primary facial features (e.g., eyes, nose, and mouth), (ii) the laplacian-of-gaussian blob detection, and (iii) morphological operators. It uses demographic information (e.g., gender and ethnicity) and facial marks (e.g., scars, moles, and freckles) for improving face image matching and retrieval performance. Josh Harguess and Aggarwal [22] explored a technique using average-half-face in 2D and 3D for face recognition. The face image is divided into half and averaged to yield the input, which is verified on six algorithms. The accuracy and potential decrease in storage and computation time are the advantages.

Piotr Porwik [23] discussed the national regulations and application requirements to introduce a new biometric passport in the European Union, especially in the republic of Poland, which is called as e-passport. The biometric systems depend highly on the existing hardware biometrical technologies and its software components. Pradeep Buddharaju et al., [24] proposed a methodology based on physiology of face in thermal infrared spectrum. The facial physiological patterns are captured using the bioheat information contained in thermal imagery. The algorithm delineates the human face from the background using the bayesian framework. Then, it localizes the superficial blood vessel network using image morphology. The extracted vascular network produces contour shapes that are characteristic to each individual.

Berkay Topcu and Hakan Erdogan [25] proposed Patch-based face recognition method, which uses the idea of analyzing face images locally, in order to reduce the effects of illumination changes and partial occlusions. Feature fusion and decision fusion are two distinct ways to utilize the extracted local features. An approach for calculating weights for the weighted sum rule is proposed. The recognition accuracy obtained with AR database is better for feature fusion methods by using validation accuracy weighting scheme and nearest-neighbor discriminant analysis dimension reduction method. Di Huang et al., [26] proposed an asymmetric 3D-2D face recognition method, enrolling people in textured 3D face model, while performing automatic identification in 2D facial images. The goal is to limit the use of 3D data where it really helps to improve face recognition accuracy. It consists of two separate matching steps: sparse representation classifier is applied to 2D-2D matching, while canonical correlation analysis is exploited to learn the mapping between LBP faces (3D) and texture LBP faces (2D). The final decision is made by combining both matching scores. A new reprocessing pipeline is used to enhance robustness to lighting and pose effects.

Archana Sapkota et al., [27] developed a novel operator called General Region Assigned to Binary (GRAB), as a generalization of LBP. They demonstrated its performance for face recognition in both constrained and unconstrained environments and across multiple scales. The GRAB significantly outperforms LBP in cases of reduced scale on subsets of two well-known published datasets of FERET and LFW. Stan Li et al., [28] developed a face recognition algorithm for indoor and co-operative user applications with variations in the illumination. The imaging system is an active near infrared camera to produce face images of good condition regardless of visible lights in the environment. The resulting face images encode intrinsic information such as 3D shape and reflectance of the facial surface which yields better results as compared to the extrinsic properties viz., eyeglasses, hairstyle, expression, posture and environmental lightening.

Zhenhua Guo et al., [29] proposed a Local Directional Derivative Pattern (LDDP) based framework, which represents higher order directional derivative features. LBPs are treated as a special case of LDDP. The higher order derivative information contains complementary features; better recognition accuracy could be achieved by combining different order LDDPs which is validated by two large public texture databases, Outex and CUReT. Hasanul Kabir et al., [30] presented a new appearance-based feature descriptor; Local Directional Pattern Variance (LDPv), to represent facial components for human expression. The LDPv introduces the local variance of directional responses to encode the contrast information within the descriptor. Here, the LDPv representation characterizes both spatial structure and contrast information of each micro-pattern. Template matching and SVM classifiers are used to classify the LDPv feature vector of different prototypic expression images.

## 3. PROPOSED DBC-FR MODEL

In this section the definitions of performance analysis and the block diagram of the proposed model are discussed.

### 3.1 Definitions

i) **False Acceptance Rate (FAR):** It is the probability that unauthorized person is incorrectly accepted as authorized person and is given by
   FAR = Number of unauthorized persons accepted / Total number of number of unauthorized persons.

ii) **False Rejection Rate (FRR):** It is the probability that authorized person is incorrectly rejected as unauthorized person and is given by
   FRR = Number of authorized persons rejected / Total number of authorized persons.

iii) **Recognition Rate (RR) :** It is the number of persons recognized correctly in the database and given by
   RR= Number of persons recognized correctly / Total number of persons in the database.

iv) **Equal Error Rate (EER) :** It is defined as the value at which both FRR and FAR are same.

### 3.2 Block Diagram of DBC-FR

The block diagram of DBC based Face Recognition using DWT (DBC-FR) to authenticate human beings is shown in Figure 1.

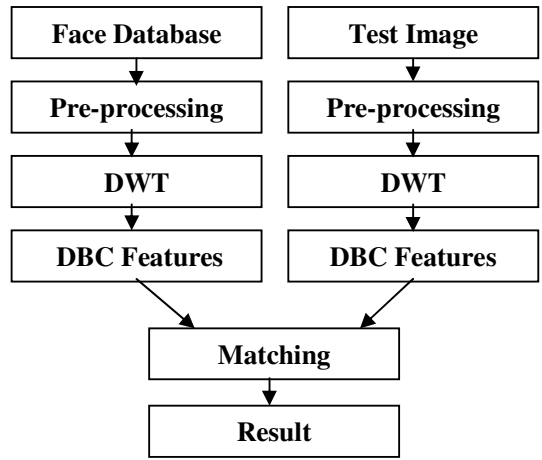

Figure 1. Block diagram of DBC-FR

### 3.2.1 Face database

The NIR spectrum is between 780 nm and 1100 nm. The poly-U NIR face database includes variation of pose, expression, Illumination, scale, blurring and combinations of them [31]. The face images of size 768*576 from NIR database are considered. The database consists of 115 persons, out of which 90 are considered for database. There are 14 images of each person, out of which 13 are considered for database and 14$^{th}$ image taken as test image. Therefore, totally there are 1170 images in the database. Figure 2 shows the fourteen samples of all NIR face images of a person. Figure 3 shows the samples of NIR face images of different persons.

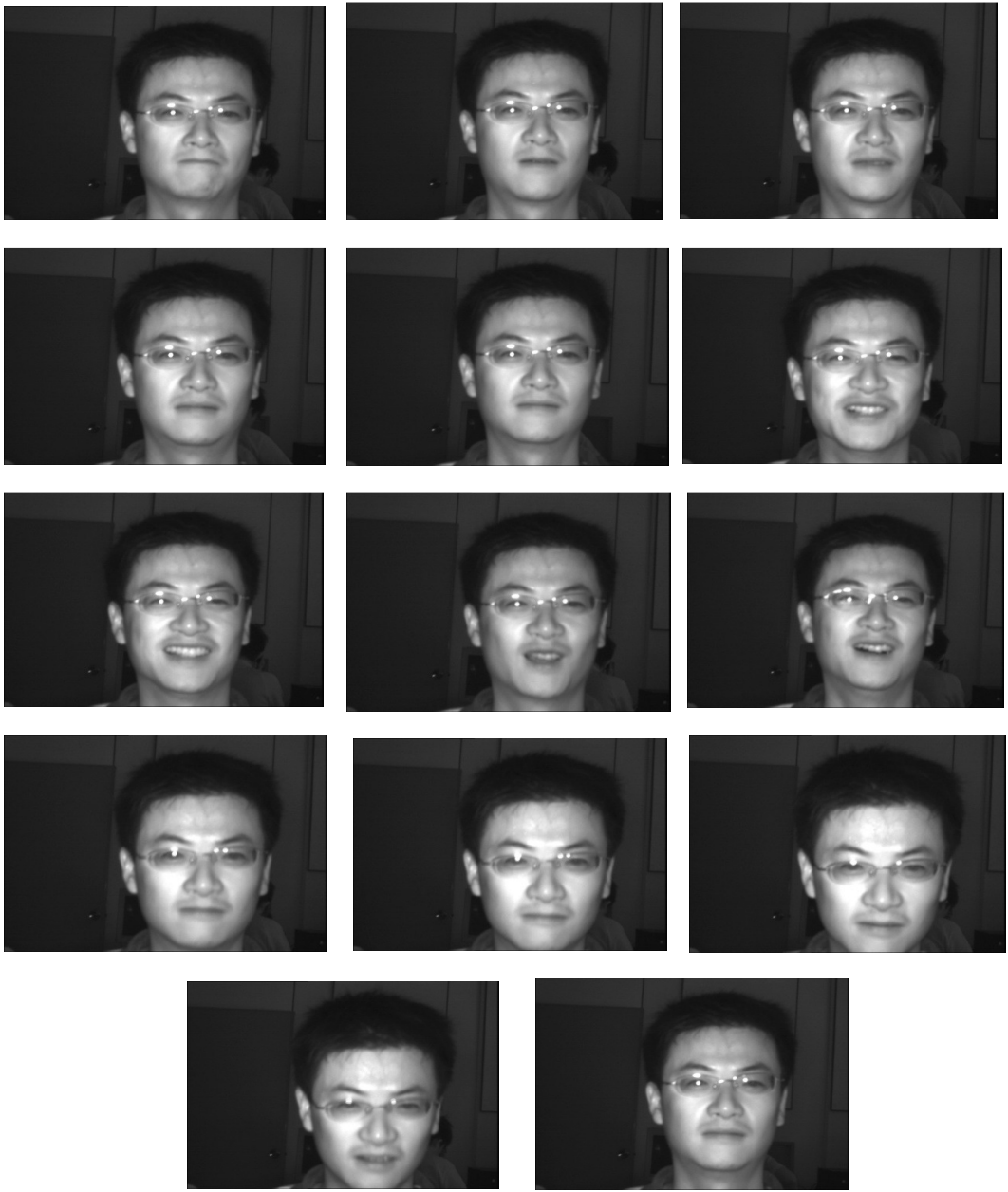

Figure 2. Fourteen face images per person of NIR database .

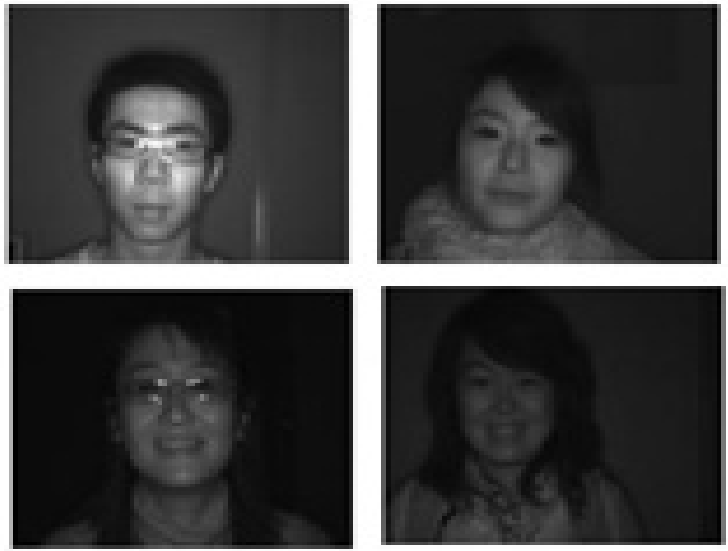

Figure 3. Samples of NIR face images of four different persons

### 3.2.2 Pre processing

The images in the database and test image are processed before extracting the features. Pre-processing involves, (i) Colour to gray scale image conversion; the gray scale image with intensity values between 0 and 255 is obtained from colour image to reduce processing time (ii) Image cropping; face image contains background and other occlusions that may not be required to identify a person correctly. Hence, only the face portion of the image is cropped. The image is converted to binary prior to cropping using the threshold value and pixel value. The binary image obtained is partitioned into two along the column. For the first portion, image is scanned from left to right until a binary 1 is encountered. When binary 1 is obtained, scanning of that row is stopped by storing the pixel's position and it is repeated for all the rows. This procedure is repeated for the second half portion scanning the image from right to left, then image is cropped based on stored pixel's positions as shown in Figure 4. (iii) Image resizing; the cropped database and test image may have different dimensions, hence the cropped image is resized to a uniform 100*100 dimension.

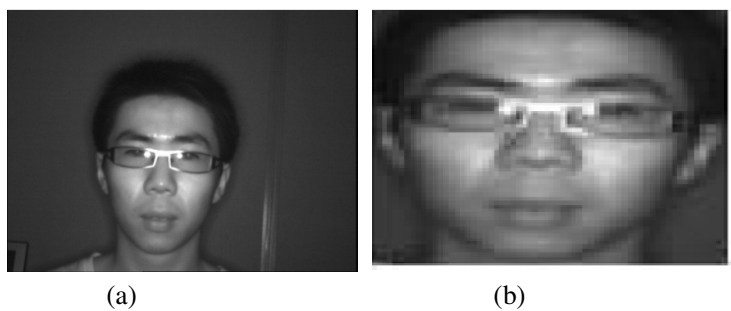

(a)          (b)

Figure 4. (a) Original image (b) Cropped image

### 3.2.3 Discrete Wavelet Transform

Wavelet transform is a powerful mathematical tool used to extract localized time-frequency (spatial-frequency) information of an image. The decomposition of the data into different frequency ranges is made using mother (prototype) wavelet and scaling functions or father wavelets and is reversible in its operation. The band pass filters perform the task of segregation of frequency components. They are classified as Continuous Wavelet Transform (CWT) and DWT; both representations are continuous in time. CWT uses all possible scales and translations, where as DWT use only selected.

DWT provides sufficient information for the analysis of original image with a significant reduction in computation time [32]. The 2D DWT is computed by successive low pass and high pass filtering the co-efficient of an image row by row and column by column. After each transform pass, the low pass co-efficient of the image may be transformed again. This process can be recursively repeated, depending the requirement. Figure 5 shows the approximation and detailed coefficients generation for the input x(n), where g(n) and h(n) are the impulse responses of low pass and high pass filters respectively for one level DWT. By applying DWT, the image is decomposed into approximation and detail components namely Low-Low (LL), High-Low (HL), Low-High (LH), High-High (HH) sub bands, corresponding to approximate, horizontal, vertical, and diagonal features respectively which are shown in Figure 6. The dimension of each sub band is half the size of original image. The sub bands HL and LH contain the changes of images or edges along horizontal and vertical directions respectively. The HH sub band contains high frequency information of the image.

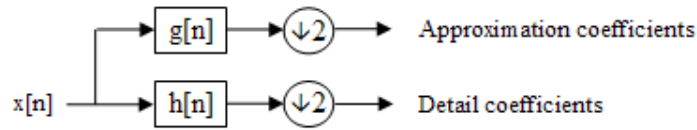

Figure 5. Block diagram of one level DWT

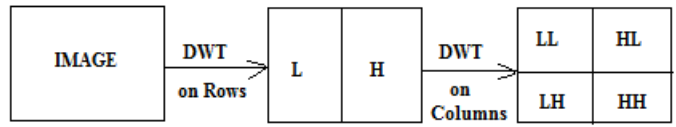

Figure 6. Four bands obtained after DWT

The 2D- DWT for an image is given in Equation 1.

$$\text{DWT}(j, k) = \frac{1}{\sqrt{2^j}} \int_{-\infty}^{\infty} f(x) \, \psi\left(\left(\frac{x}{2}\right) - k\right) dx \quad (1)$$

where *j* is the power of binary scaling and *k* is a constant of the filter. One- level DWT is applied on each face image. Haar wavelet is used as the mother wavelet, since it is simple, faster, reversible, memory efficient and gives better results compared to other wavelets. Approximate band ie., LL band is considered as the most significant information of the face image. The size of the image obtained after DWT is 50*50. The Haar wavelet is a sequence of rescaled square-shaped functions which together form a wavelet family or basis. The basic functions involved are; averaging and differencing the input data by many levels yielding coefficients. The Haar wavelet's function is given by the Equation 2 and its scaling function [33] is given by the Equation 3.

$$\psi(t) = \begin{cases} 1, & 0 \leq t < 1/2 \\ -1, & 1/2 \leq t < 1 \\ 0, & \text{otherwise} \end{cases} \qquad (2)$$

$$\phi(t) = \begin{cases} 1, & 0 \leq t < 1 \\ 0, & \text{otherwise} \end{cases} \qquad (3)$$

### 3.2.4 DBC Features

The process of face recognition involves identifying the person based on facial features. Hence it is necessary to extract the features from a given image. DBC is applied on LL sub band to encode the directional edge information. It captures the spatial relationship between any pair of neighbourhood pixels in a local region along a given direction. It reflects the image local feature. It extracts more spatial information than LBP. Let $Z_{i,j}$ be a point in a cell, four directional derivatives at $Z_{i,j}$ are given in Equation 4 to Equation 7.

$$I'_{0,d}(z_{i,j}) = I(z_{i,j}) - I(z_{i,j-d}) \qquad (4)$$
$$I'_{45,d}(z_{i,j}) = I(z_{i,j}) - I(z_{i-d,j+d}) \qquad (5)$$
$$I'_{90,d}(z_{i,j}) = I(z_{i,j}) - I(z_{i-d,j}) \qquad (6)$$
$$I'_{135,d}(z_{i,j}) = I(z_{i,j}) - I(z_{i-d,j-d}) \qquad (7)$$

The resized image of size 50*50 is partitioned into 100 cells of 5*5 matrixes. Table 1 shows a 3*3 neighbourhood centre on $I_{i,j}$ taken out of 5*5 cell size, where each cell contributes one coefficient. For each cell, first order derivatives, denoted as $I'_{\alpha,d}$ is calculated, where $\alpha = 0^0, 45^0, 90^0, 135^0$ and d is the distance between the given point and its neighbouring point. Each coefficient values along four directions are averaged respectively to yield hundred features per face image. The derivatives obtained are converted into binary using the Equation 8.

Table 1. A 3*3 neighbourhood centre on $I_{i,j}$

| $I_{i-1,j-1}$ | $I_{i,j-1}$ | $I_{i+1,j-1}$ |
|---|---|---|
| $I_{i-1,j}$ | $I_{i,j}$ | $I_{i+1,j}$ |
| $I_{i-1,j+1}$ | $I_{i,j+1}$ | $I_{i+1,j+1}$ |

$$f(I'_{\alpha,d}(z)) = \begin{cases} 0, & I'_{\alpha,d}(z) \leq 0 \\ 1, & I'_{\alpha,d}(z) > 0 \end{cases} \qquad (8)$$

Out of 25 values of a 5*5 matrix, only centre 9 values are extracted to form a 9 bit binary code. Those 9 bits are read according to the Equation 9. The DBCs computed along four directions ie., $0^0, 45^0, 90^0$ and $135^0$ are as shown in Figure 7, given by: $DBC_{0,1}$ = 010000010, $DBC_{45,1}$ = 111101010, $DBC_{90,1}$ = 111011010, $DBC_{135,1}$ = 001101000. The Figure 8 shows the respective DBC feature maps of an image.

$$DBC_{\alpha,d}(z_{x,y}) = \{ f(I'_{\alpha,d}(z_{x,y})) ; f(I'_{\alpha,d}(z_{x,y-d})) ; f(I'_{\alpha,d}(z_{x-d,y-d})) ;$$
$$f(I'_{\alpha,d}(z_{x-d,y})) ; f(I'_{\alpha,d}(z_{x-d,y+d})) ; f(I'_{\alpha,d}(z_{x,y+d})) ;$$
$$f(I'_{\alpha,d}(z_{x+d,y+d})) ; f(I'_{\alpha,d}(z_{x+d,y})) ; f(I'_{\alpha,d}(z_{x+d,y-d})) \} \qquad (9)$$

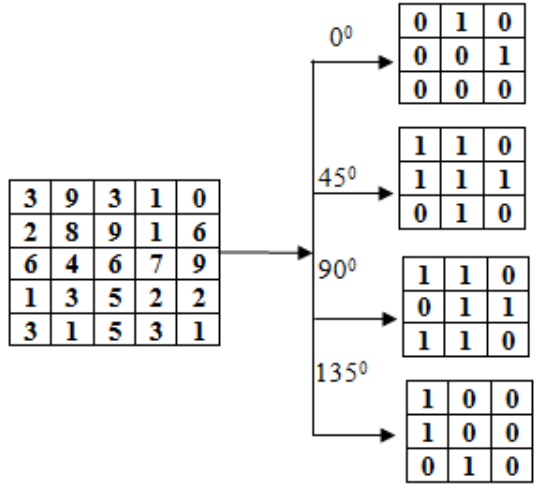

Figure 7. DBCs along $0^0$, $45^0$, $90^0$ and $135^0$ respectively

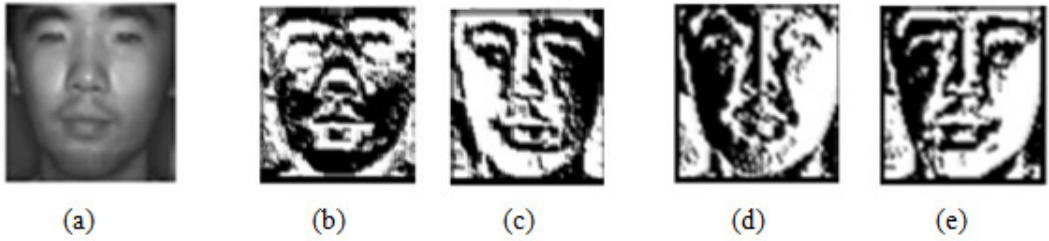

Figure 8. (a) Original image, (b) to (e) DBC feature maps along $0^0$, $45^0$, $90^0$ and $135^0$ -respectively.

### 3.2.5 Matching

Euclidean Distance (ED) is used to verify whether the person is in database or not. If $p_i$ and $q_i$ are two points in a 2D plane, where i =1, 2, then the ED is given by the Equation 10.

$$d(p,q) = \sqrt{((p_1-q_1)^2 + (p_2-q_2)^2)} \qquad (10)$$

The ED is calculated by comparing the feature vector of one test image with feature vectors of all images in the database. ED value and position of the image in the database for which ED is minimum and Person number is noted. ED value is compared with the threshold value. Here the range of threshold is from 0 to 1.2. If the ED value is less than the threshold, we have to check whether the person from the database and test image of a person is same. If it is same, then the match count is incremented, else the mismatch count is incremented. If the ED value is greater than threshold then the false rejection rate count is incremented indicating the image in database is falsely rejected.

Now the test image is taken out of the database and above procedure is repeated to calculate the ED value. Then if ED value is greater than threshold, mismatch count is incremented. If ED value is less than threshold, false acceptance count is incremented, indicating the image is not present in database but still matching with a person from the database.

## 4. ALGORITHM

Problem definition: The DBC based Face Recognition using DWT is used to identify a person. The objectives are

  i. To increase the RR
 ii. To decrease FRR and FAR

The algorithm of the proposed DBC-FR model is shown in Table 2.

Table 2. Algorithm of DBC-FR

> **Input: Face Database, Test Face image**
> **Output: Recognition/ Rejection of a person.**
> 1. Image is preprocessed.
> 2. DWT is applied and approximate band is considered.
> 3. LL band of 50*50 is partitioned into 100 cells of 5*5 matrixes each.
> 4. For each cell, the directional derivatives along $0^0$, $45^0$, $90^0$ and $135^0$ are computed.
> 5. Derivatives are converted into binary and read in anticlockwise direction to form 9-bit binary code.
> 6. 9-bit code generated is then converted into its decimal equivalent.
> 7. The decimal values of all directional derivatives of each cell are averaged to obtain single co-efficient per cell and 100 coefficients for whole image constitutes features.
> 8. Euclidean distance between feature vectors of images in database and feature vectors of test image is computed.
> 9. Image with minimum Euclidean distance is considered as matched image.

## 5. PERFORMANCE ANALYSIS

The Table 3 shows the variations of performance parameters such as RR, FAR and FRR for different values of threshold. As the value of threshold increases, the values of FRR and FAR decreases and increases respectively. The RR is zero till 0.4 threshold value, from threshold value 0.6, the RR value increases from 13.3% to 98.89% for the threshold value 1.2. The percentage RR for an existing algorithm *Directional Binary Code with Application to poly-U Near-Infrared Face Database (Gabor-DBC)* [6] and the proposed DBCFR algorithm is compared in Table 4. The DBC applied on LL sub band of DWT to extract features improves the % RR in the proposed algorithm compared to the existing algorithm. The figure 9 depicts the variation of FRR and FAR with threshold and it is noted that the value of EER is 0.8214 for 0.6214 value of threshold i.e. at 0.6214 threshold the value of FRR and FAR are equal.

Table 3. Variation of FRR, FAR and RR with Threshold

| Threshold | FRR | FAR | % RR |
|---|---|---|---|
| 0 | 1 | 0 | 0 |
| 0.2 | 1 | 0 | 0 |
| 0.4 | 1 | 0 | 0 |
| 0.6 | 0.8667 | 0.8 | 13.3 |
| 0.8 | 0.4444 | 1 | 55.6 |
| 1.0 | 0.1000 | 1 | 90.00 |
| 1.2 | 0.0111 | 1 | 98.89 |

Table 4. Comparison of % RR

| Parameter | Existing Gabor-DBC [6] | Proposed DBC-FR |
|---|---|---|
| % RR | 97.6 | 98.8 |

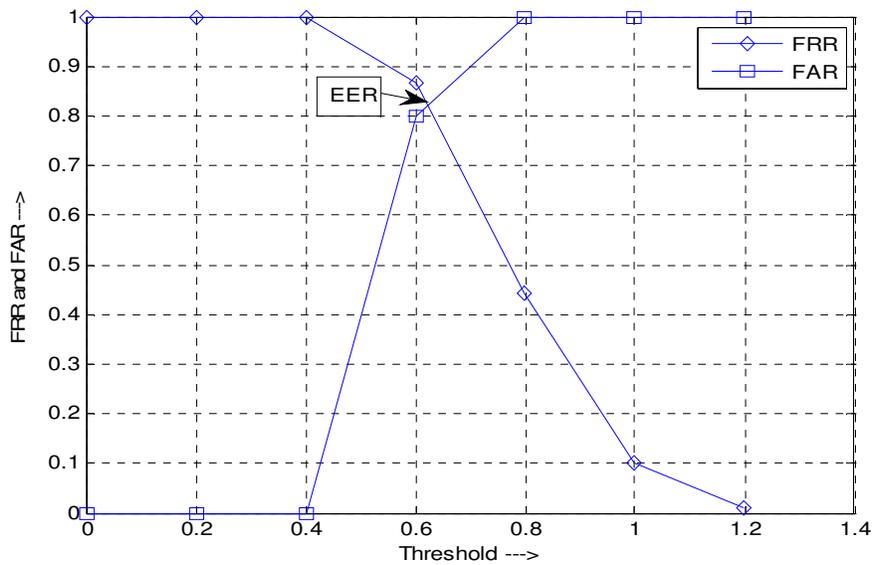

Figure 9. Variation of FRR and FAR with Threshold

## 6. CONCLUSION

The face image is physiological trait and is a better biometric data as the samples can be obtained without the co-operation of a person as well as can also be captured with reasonable distance. The Poly-U near infra red face database is used to evaluate the performance of an algorithm. In this paper DBC-FR algorithm is proposed. The Face images are pre-processed to get only face part using scanning and cropping. The face image is resized to 100*100 dimension. The DWT is applied and only LL sub band of size 50 *50 is considered. The LL sub band is segmented into one hundred cells each of size 5*5. The DBC is applied on each cell to constitute 100 features. The Euclidean distance is used for comparison. The values of RR, FAR and FRR are computed. It is observed that the value of RR is improved in the proposed algorithm compared to the existing algorithm. In future the performance of an algorithm can be improved by the fusion of spatial and transform domain techniques.

## ACKNOWLEDGEMENT

I would like to thank Department of Electronics and Communication Engineering, Rayalaseema University, Kurnool - 518002, Andhra Pradesh, India for allowing me to pursue research work.

## Authors

**Jagadeesh H S** awarded the B.E degree in E & C Engineering from Bangalore University and M.Tech degree in Digital Electronics and Communication Systems from Visvesvaraya Technological University, Belgaum. He is pursuing his Ph.D. in Electronics and Communication Engineering of Rayalaseema University, Kurnool under the guidance of Dr. K. Suresh Babu, Assistant Professor, Department of Electronics and Communication Engineering. He has one research publication in refereed International Conference Proceedings. He is currently an Assistant Professor, Dept. of Electronics and Communication Engineering, A P S College of Engineering, Bangalore. His research interests include Image processing, Pattern Recognition, and Biometrics. He is a student member of IEEE and life member of Indian Society for Technical Education, New Delhi.

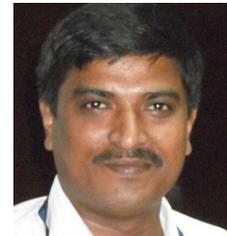

**K. Suresh Babu** is an Assistant Professor, Dept. of Electronics and Communication Engineering, University Visvesvaraya College of Engineering, Bangalore University, Bangalore. He obtained his BE and ME in Electronics and Communication Engineering from University Visvesvaraya College of Engineering Bangalore. He was awarded Ph.D. in Computer Science and Engineering from Bangalore University. He has over 15 research publications in refereed International Journals and Conference Proceedings. His research interests include Image Processing, Biometrics, Signal Processing

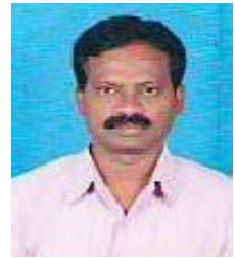

**K B Raja** is an Assistant Professor, Dept. of Electronics and Communication Engineering, University Visvesvaraya College of Engineering, Bangalore University, Bangalore. He obtained his BE and ME in Electronics and Communication Engineering from University Visvesvaraya College of Engineering Bangalore. He was awarded Ph.D. in Computer Science and Engineering from Bangalore University. He has over 90 research publications in refereed International Journals and Conference Proceedings. His research interests include Image Processing, Biometrics, VLSI Signal Processing, Computer Networks.

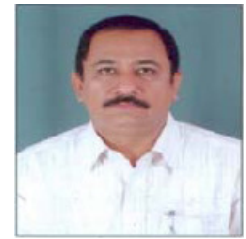